\documentclass[10pt,twocolumn,letterpaper]{article}

\usepackage{wacv}
\usepackage{times}
\usepackage{epsfig}
\usepackage{graphicx}
\usepackage{amsmath}
\usepackage{amssymb}
\usepackage{algorithm}
\usepackage{algorithmic}
\usepackage{adjustbox}
\usepackage{multirow}
\usepackage{float}

\usepackage[maxcitenames = 1]{biblatex}
\author{Rishabh Iyer \\ Microsoft Corporation \\ {\tt\small rishi@microsoft.com}
\and Pratik Dubal \\ AitoeLabs \\ {\tt\small pratik@aitoelabs.com} \and Kunal Dargan \\ AitoeLabs \\ {\tt\small kunal@aitoelabs.com
}
\and Suraj Kothawade \\ IIT Bombay \\ {\tt\small surajkothawade@cse.iitb.ac.in
} \and Rohan Mahadev \\ AitoeLabs \\ {\tt\small rohan@aitoelabs.com} \and Vishal Kaushal \\ IIT Bombay \\ {\tt\small vkaushal@cse.iitb.ac.in}}

\def\wacvPaperID{589} 

\wacvfinalcopy

\ifwacvfinal\fi
\begin{document}
\title{Vis-DSS: An Open-Source toolkit for Visual Data Selection and Summarization}
\maketitle
\ifwacvfinal\thispagestyle{empty}\fi
\begin{abstract}
With increasing amounts of visual data being created in the form of videos and images, visual data selection and summarization are becoming ever increasing problems. We present Vis-DSS, an open-source toolkit for Visual Data Selection and Summarization. Vis-DSS implements a framework of models for summarization and data subset selection using submodular functions, which are becoming increasingly popular today for these problems. We present several classes of models, capturing notions of diversity, coverage, representation and importance, along with optimization/inference and learning algorithms. Vis-DSS is the first open source toolkit for several Data selection and summarization tasks including Image Collection Summarization, Video Summarization, Training Data selection for Classification and Diversified Active Learning. We demonstrate state-of-the art performance on all these tasks, and also show how we can scale to large problems. Vis-DSS allows easy integration for applications to be built on it, also can serve as a general skeleton that can be extended to several use cases, including video and image sharing platforms for creating GIFs, image montage creation, or as a component to surveillance systems and we demonstrate this by providing a graphical user-interface (GUI) desktop app built over Qt framework. Vis-DSS is available at \url{https://github.com/rishabhk108/vis-dss}. \looseness-1 
\end{abstract}
\section{Introduction} \label{introduction}
Visual Data in the form of images, videos and live streams have been growing in the last few years. While this massive data is a blessing to data science by helping improve predictive accuracy, it is also a curse since humans are unable to consume this large amount of data. Drones, Dash-cams, Body-cams, Security cameras, Go-pro and Cell phones etc. generate videos and photos at a rate higher than what we can process. Moreover, majority of this data is plagued with redundancy. Given this data explosion, machine learning techniques which automatically understand, organize and categorize this data are of utmost importance.

Visual Data summarization attempts to provide a highlight of the most critical and important events in the video, giving the viewer a quick glimpse of the entire video so they can decide which parts of the video is important. Similarly Machine Learning models today require training data often labeled by humans. Visual Data Subset Selection attempts to summarize large training data-sets saving both the time and the cost of human labeling efforts. 

We present Vis-DSS, a toolkit for both Visual Data Subset Selection for Training Vision Models, and for summarization of videos and image collections. Our engine uses Submodular Function optimization, and presents several models of diversity, coverage, representation etc. We also empirically compare these different models and suggest recipes for using these models in different scenarios.

In order to perform summarization, our system accepts input in the form of an image collection/video, analyses and pre-processes it to generate meta-information about the input. The meta-information consists of visual features, scene information, presence of objects, faces and colors. This meta-information, along with information such as the desired size of the summary output, form the input to our summarization module.

Since modern deep learning models require quite a lot of data, sourcing the data and labeling them is a time consuming and expensive ordeal. There are also other problems with visual training data like data redundancy, where the amount of new information we add to the model is not proportionate to the amount of training data added. This increases the cost of labeling the data without much utility. Hence, selecting a rich subset from an existing training set without losing out on too much accuracy becomes an important tool to quickly train deep models. Summarization techniques can be used to choose such a rich subset from the visual training data to help aid in this process.

Learning approaches like active learning, which is contradictory to passive learning (supervised learning), gives the algorithm a privilege to select the best samples for optimum learning. However, making the right choice of data can be a cumbersome task in itself. Submodular functions, if used in the right way, can be a great query selection strategy in active learning that help in selecting best samples in the correct amount.  

Vis-DSS provides a unified software framework with several applications including a) Image collection summarization, b) Video Summarization, c) Visual Training data subset selection, and d) Diversified active learning. All these applications are built over our core engine of submodular function optimization.

\section{Related Work and Our Contributions}
There have been various approaches in the past to solve Image Collection Summarization and Video Summarization. \cite{yang2013image} perform Image Collection Summarization by formulating a summarization problem as a dictionary learning problem. Other approaches, such as \cite{tschiatschek2014learning, sceneSummImage}, tackle the problem by identifying the most representative set of images in the collection. \citeauthor{tschiatschek2014learning} \cite{tschiatschek2014learning} tackled the Image Collection Summarization problem using submodular functions. They argue how a lot of previous work on this problem, have, often unknowingly, used models which are submodular. They unify all this work by proposing a framework for learning mixtures of submodular functions for image collection summarization, and also propose an evaluation measure for this problem.

Video Summarization approaches differ in the representation of the output summary. Some works extract key-frames \cite{wolfKeyFrame, egoSum, khoslaSum, kimSum} from the input video, while some generate skims \cite{gyglim, gygli2014creating} from it. There have also been other forms of video summarization which include creation of montages \cite{rankHighlights} and GIFs \cite{vid2gif}. Building upon existing work from~\cite{tschiatschek2014learning, lin2011class}. \citeauthor{gyglim} \cite{gyglim} also utilized submodular functions for performing video summarization. They also learn a mixture of submodular functions to generate summaries by providing a trade-off for interestingness, representation and uniformity. 
Related approaches to visual training data subset selection has revolved around the idea of choosing a smaller subset which is characteristically more diversity or representative of the overall dataset. As submodular functions are able to mathematically model these ideas of diversity and representativeness of a set, they are put to use in data subset selection applications for computer vision tasks such as text categorization \cite{pmlr-v37-wei15} as well as for speech recognition. \cite{DBLP:conf/icassp/WeiLKBB14}

In the light of learning from lesser data, researchers have looked into techniques for selecting the right samples for training. This technique is famously known as active learning. There are limited studies that amalgamate active learning and subset selection for computer vision problems. \citeauthor{sourati2016classification} used information theoretic objective function like mutual information between the labels and developed a framework \cite{sourati2016classification} to evaluate and maximize this objective for batch mode active learning. Another open-source toolkit Jensen \cite{jensen} has support for continuous convex optimization which is essential for active learning aspects.  \citeauthor{ducoffescalable} have incorporated batch active learning to deep models. It is common in active learning to use a core-set \cite{agarwal2005geometric} which efficiently approximate various geometric extent measures for large set of data instances by using a smaller, but diverse subset.

There is somewhat limited work around open source software and systems for data summarization. \citeauthor{gyglim} \cite{gyglim} have also released their code for video summarization, along with a dataset for evaluation. Similarly \citeauthor{krause2010sfo} \cite{krause2010sfo} has previously released a toolbox for Submodular Function Optimization.  However, both these do not adapt well to the needs of large scale problems. 
This paper releases a first of its kind, Visual Data Selection and Summarization toolkit, which uses Submodular Functions Optimization to identify diverse and representative elements in a set, while minimizing redundancy. We provide several classes of submodular functions including Facility Location, Minimum Dispersion, Graph Cut, Saturated Coverage, Set Cover, Probabilistic Set Cover, Feature Based functions etc. These functions have extensively been used in several summarization papers~\cite{lin2011class, tschiatschek2014learning, gyglim, wei2015submodularity, wei2014fast, sahoo2017unified}. We also provide three variants of inference/optimization algorithms including Greedy algorithms for Submodular Knapsack problem~\cite{sviridenko2004note}, Submodular Cover~\cite{wolsey1982analysis} and a Streaming Greedy algorithm. In each case, we provide Lazy greedy implementations~\cite{minoux1978accelerated}. Finally, and most importantly, we provide several implementational tricks including a Memoization framework. We show how using Memoization, we are able to achieve significant speedups to optimization/inference algorithms. We note that none of the existing softwares for submodular optimization~\cite{krause2010sfo, gyglim} implement these memoization tricks.

Building upon our engine, we provide several applications including Image collection summarization, Video summarization, Data subset selection and active learning. We provide three kinds of video/image summarization, including extractive summarization, query based and entity summarization. We support several models for feature extraction, including simple color histogram based features right upto deep model based feature extractors. We finally evaluate our framework on several datasets and demonstrate how our toolkit is capable of state-of-the art performance for these problems.

\begin{figure*}
\begin{center}
\includegraphics[width = 0.3\textwidth]{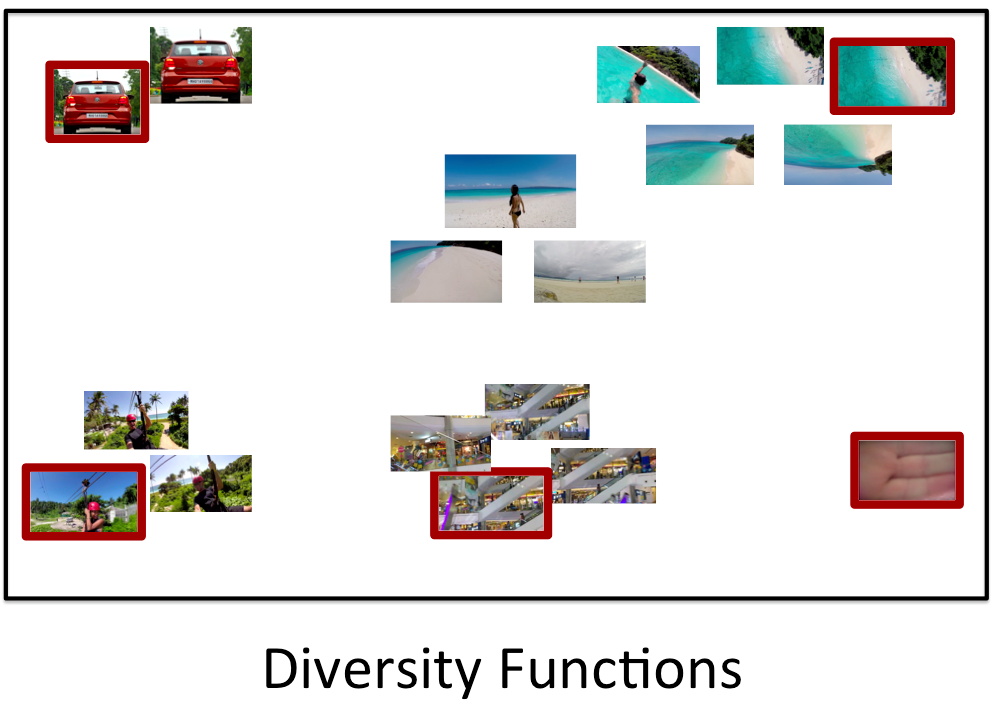}
\includegraphics[width = 0.3\textwidth]{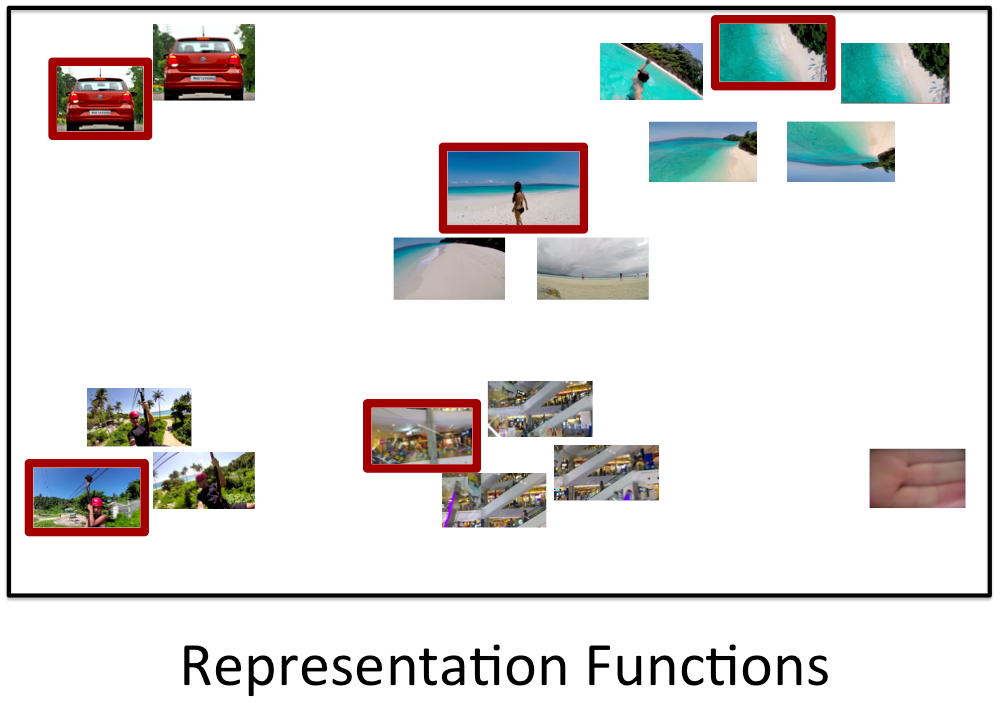}
\includegraphics[width = 0.3\textwidth]{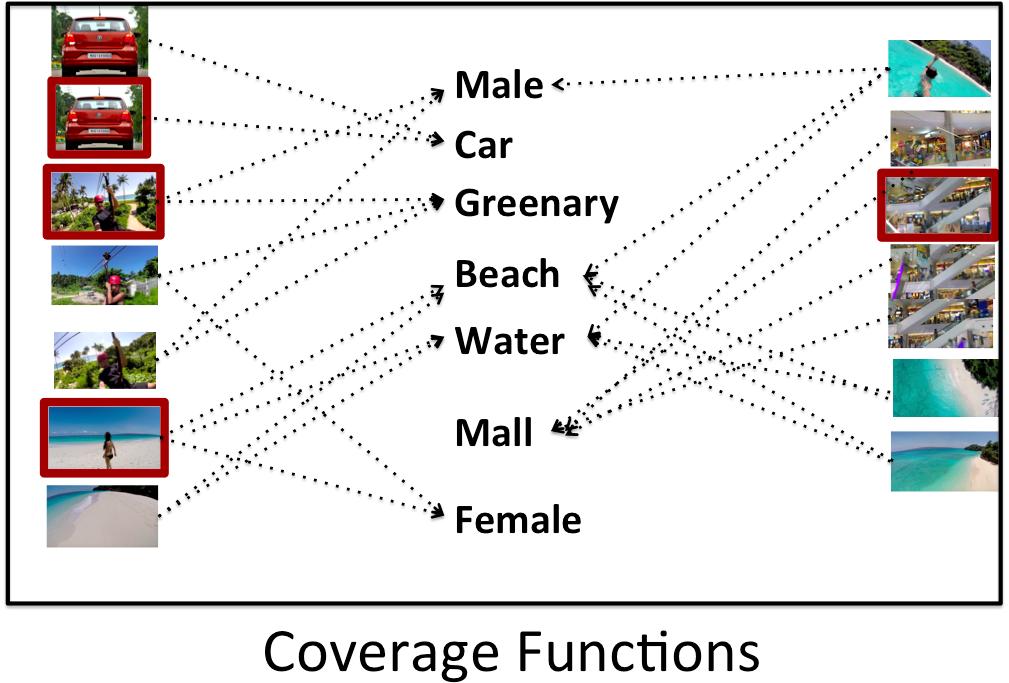}
\end{center}
\caption{Illustration of the Difference between Diversity Functions, Coverage Functions and Representation Functions}
\label{divrepcov}
\end{figure*}

\section{Submodular Summarization Models}
This section describes the Submodular Functions used as Summarization Models in our system. We divide these into Coverage Functions, Representation Functions and Diversity Functions.

\subsection{Modeling Coverage}
This class of functions model notions of coverage, i.e. try to find a subset of the ground set $X$ which covers a set of \emph{concepts}. Below are instantiations of this.

\subsubsection{Set Cover Function} Denote $V$ as the ground set and let $X \subseteq V$ be a subset (of snippets or frames). Further $\mathcal U$ denotes a set of concepts, which could represent, for example, scenes or objects. Each frame (or snippet) $i \in X$ contains a subset $U_i \in \mathcal U$ set of concepts (for example, an image covers a table, chair and person). The set cover function then is 
\begin{align}
f(X) = w(\cup_{i \in X} U_i), 
\end{align}
where $w_u$ denotes the weight of concept $u$. 

\subsubsection{Probabilistic Set Cover} This is a generalization of the set cover function, to include probabilities $p_{i u_i}$ for each object $u_i$ in Image $i \in X$. For example, our Convolutional Neural Network (CNN) might output a confidence of object $u_i$ in Image $i$, and we can use that in our function. The probabilistic coverage function is defined as, 
\begin{align}
f(X) = \sum_{i \in \mathcal U} w_i[1 - \prod_{i \in X}(1 - p_{ij})].
\end{align}
The set cover function is a special case of this if $p_{ij} = 1$ if Object $j$ belongs to Image $i$ (i.e. we use the hard labels instead of probabilities).

\subsubsection{Feature Based Functions} Finally, we investigate the class of Feature Based functions. Here, we denote an Image $i$ via a feature representation $q_i$. This could be, for example, the features extracted from a fully connected layer of a CNN. Denote $F$ as the set of features. The feature based function is defined as,
\begin{align}
f(X) = \sum_{i \in F} \psi(q_i(X))
\end{align}
where $q_i(X) = \sum_{j \in X} q_{ij}$, and $q_{ij}$ is the value of feature $i$ in Image $j$. $\psi$ is a concave function.  Examples of $\psi$ are square-root, Log and Inverse Function etc.

\subsection{Modeling Representation}
Representation based functions attempt to directly model representation, in that they try to find a representative subset of items, akin to centroids and mediods in clustering.

\subsubsection{Facility Location Function} The Facility Location function is closely related to k-mediod clustering. Denote $s_{ij}$ as the similarity between images $i$ and $j$. We can then define $f(X) = \sum_{i \in V} \max_{j \in X} s_{ij}$. For each image $i$, we compute the representative from $X$ which is closest to $i$ and add the similarities for all images. Note that this function, requires computing a $O(n^2)$ similarity function. However, as shown in~\cite{wei2014fast}, we can approximate this with a nearest neighbor graph, which will require much smaller space requirement, and also can run much faster for large ground set sizes.

\subsubsection{Saturated Coverage Function} The saturated coverage function~\cite{lin2011class} is defined as $f(X) = \min\{\sum_{i \in X} s_{ij}, \alpha \sum_{i \in V} s_{ij}\}$. This function is similar to Facility Location and attempts to model representation. This is also a Kernel based function and requires computing a similarity matrix.

\subsubsection{Graph Cut Functions} We define the graph cut family of functions as $f(X) = \lambda \sum_{i \in V} \sum_{j \in X} s_{ij} - \sum_{i, j \in X} s_{ij}$. This function is similar to the Facility Location and Saturated Coverage in terms of its modeling behaviour.

\subsection{Modeling Diversity}
The third class of Functions are Diversity based ones, which attempt to obtain a diverse set of key points. 

\subsubsection{Dispersion (Disparity) Functions} Denote $d_{ij}$ as a distance measure between Images $i$ and $j$. Define a set function $f(X) = \min_{i, j \in X} d_{ij}$. This function is not submodular, but can be efficiently optimized via a greedy algorithm~\cite{dasgupta2013summarization}. It is easy to see that maximizing this function involves obtaining a subset with maximal minimum pairwise distance, thereby ensuring a diverse subset of snippets or key-frames. Similar to the Minimum Disparity, we can define two more variants. One is Disparity Sum, which can be defined as $f(X) = \sum_{i, j \in X} d_{ij}$. This is a supermodular function. Another model is, what we call, Disparity Min-Sum which is a combination of the two forms of models. Define this as $f(X) = \sum_{i \in X} \min_{j \in X} d_{ij}$. This function is submodular~\cite{chakraborty2015adaptive}.

\subsection{Understanding Diversity, Representation and Coverage}
Figure~\ref{divrepcov} demonstrates the intuition of using diversity, representation and coverage functions. Diversity based functions attempt to find the most different set of images. The leftmost figure demonstrates this. It is easy to see that the five most diverse images are picked up by the diversity function (Disparity Min), and moreover, the summary also contains the image with a hand covering the camera (the image on the right hand side bottom), which is an outlier. The middle figure demonstrates the summary obtained via a representation function (like Facility Location). The summary does not include outliers, but rather contains one representative image from each cluster. The diversity function on the other hand, does not try to achieve representation from every cluster. The rightmost figure demonstrates coverage functions. The summary obtained via a coverage function (like Set Cover or Feature based function), covers all the concepts contained in the images (Male, Car, Greenery, Beach etc.). 

\subsection{Optimization/Inference Algorithms}
In the previous section, we discussed the different submodular functions used in our summarization model. We now investigate the various optimization algorithms used by us to solve our problems.

\subsubsection{Budget Constrained Submodular Maximization} For Budget Constrained Submodular Maximization, the greedy algorithm is a slight variant, where at every iteration, we sequentially update $X^{t+1} = X^t \cup \mbox{argmax}_{j \in V \backslash X^t} \frac{f(j | X^t)}{c(j)}$. This algorithm has near optimal guarantees~\cite{sviridenko2004note}. 

\subsubsection{Submodular Cover Problem} For the Submodular Cover Problem, we again resort to a greedy procedure~\cite{wolsey1982analysis} which is near optimal. In this case, the update is similar to that of problem 1, i.e. choose $X^{t+1} = X^t \cup \mbox{argmax}_{j \in V \backslash X^t} f(j | X^t)$. We stop as soon as $f(X^t) = f(V)$, or in other words, we achieve a set which covers all the concepts.

\subsubsection{Stream Greedy} Finally, we describe the Stream greedy algorithm. We implement a very simple version of streaming greedy algorithm. We first order the elements via a permutation $\sigma$. Denote $\sigma[i]$ as the $i$ element in the ordering. Given a threshold $\tau$, the streaming greedy ads an element to the summary if the gain of adding element $\sigma[i]$ to the current summary $X$ is greater than $\tau$, i.e. if $f(\sigma[i] | X) \geq \tau$. 

\begin{table*}
\begin{center}
 \begin{tabular}{|| c | c | c |  c | c ||} 
 \hline
 Name & $f(X)$ & $p_f(X)$ & $T^o_f$ & $T^p_f$\\ [0.5ex] 
 \hline
 Facility Location & $\sum_{i \in V} \max_{k \in X} s_{ik}$ & $[\max_{k \in X} s_{ik}, i \in V]$ & $O(n^2)$ & $O(n)$\\ 
 \hline
 Saturated Coverage & $\sum_{i \in V} \min\{\sum_{j \in X} s_{ij}, \alpha_i\}$ & $[\sum_{j \in X} s_{ij}, i \in V]$ & $O(n^2)$ & $O(n)$\\ 
\hline
 Graph Cut & $\lambda \sum_{i \in V}\sum_{j \in X} s_{ij} - \sum_{i, j \in X} s_{ij}$ & $[\sum_{j \in X} s_{ij}, i \in V]$ & $O(n^2)$ & $O(n)$\\ 
\hline
 Feature Based & $\sum_{i \in \mathcal F} \psi(w_i(X))$ & $[w_i(X), i \in \mathcal F]$ & $O(n|\mathcal F|)$ & $O(|\mathcal F|)$ \\
 \hline
 Set Cover & $w(\cup_{i \in X} U_i)$ & $\cup_{i \in X} U_i$ & $O(n|U|$ & $|U|$\\
 \hline
 Prob. Set Cover & $\sum_{i \in \mathcal U} w_i[1 - \prod_{k \in X}(1 - p_{ik})]$ & $[\prod_{k \in X} (1 - p_{ik}), i \in \mathcal U]$ & $O(n|\mathcal U|)$ & $O(|\mathcal U|)$ \\ [1ex] 
 \hline
 Dispersion Min & $\min_{k,l  \in X, k \neq l} d_{kl}$ & $\min_{k, l \in X, k \neq l} d_{kl}$ & $O(|X|^2)$ & $O(|X|)$\\
 \hline
 Dispersion Sum & $\sum_{k,l  \in X} d_{kl}$ & $[\sum_{k \in X} d_{kl}, l \in X]$ & $O(|X|^2)$ & $O(|X|)$\\
  \hline
 Dispersion Min-Sum& $\sum_{k \in X} \min_{l \in X} d_{kl}$ & $[\min_{k \in X} d_{kl}, l \in X]$ & $O(|X|^2)$ & $O(|X|)$\\
 \hline
\end{tabular}
\caption{List of Submodular Functions used, with the precompute statistics $p_f(X)$, gain evaluated using the precomputed statistics $p_f(X)$ and finally $T^f_o$ as the cost of evaluation the function without memoization and $T^f_p$ as the cost with memoization. It is easy to see that memoization saves an order of magnitude in computation.}
\end{center}
\end{table*}

\subsubsection{Lazy Greedy Implementations} Each of the greedy algorithms above admit lazy versions which run much faster than the worst case complexity above~\cite{minoux1978accelerated}. The idea is that instead of recomputing $f(j | X^t), \forall j \notin ^t$, we maintain a priority queue of sorted gains $\rho(j), \forall j \in V$. Initially $\rho(j)$ is set to $f(j), \forall j \in V$. The algorithm selects an element $j \notin X^t$, if $\rho(j) \geq f(j | X^t)$, we add $j$ to $X^t$ (thanks to submodularity). If $\rho(j) \leq f(j | X^t)$, we update $\rho(j)$ to $f(j | X^t)$ and re-sort the priority queue. The complexity of this algorithm is roughly  $O(k n_R T_f)$, where $n_R$ is the average number of re-sorts in each iteration. Note that $n_R \leq n$, while in practice, it is a constant thus offering almost a factor $n$ speedup compared to the simple greedy algorithm.

\subsubsection{Implementation Tricks: Memoization}
This section goes over implementation tricks via memoization. One of the parameters in the lazy greedy algorithms is $T_f$, which involves evaluating $f(X \cup j) - f(X)$. One option is to do a na\"{\i}ve implementation of computing $f(X \cup j)$ and then $f(X)$ and take the difference. However, due to the greedy nature of algorithms, we can use memoization and maintain a precompute statistics $p_f(X)$ at a set $X$, using which the gain can be evaluated much more efficiently. At every iteration, we evaluate $f(j | X)$ using $p_f(X)$, which we call $f(j | X, p_f)$.  We then update $p_f(X \cup j)$ after adding element $j$ to $X$. Table 1 provides the precompute statistics, as well as the computational gain for each choice of a submodular function $f$. Denote $T_f^o$ as the time taken to na\"{\i}vely compute $f(j | X) = f(X \cup j) - f(X)$. Denote $T_o^p$ as the time taken to evaluate this gain given the pre-compute statistics $p_X$. We see from Table 1, that evaluating the gains using memoization is often an order of magnitude faster. Moreover, notice that we also need to update the pre-compute statistics $p_X$ at every iteration. For the functions listed in Table 1, the cost of updating the pre-compute statistics is also $T_f^p$. Hence every iteration of the (lazy) greedy algorithm costs only $2T_f^p$ instead of $T_f^o$ which is an order of magnitude larger in every case. In our results section, we evaluate empirically the benefit of memoization in practice.

\section{Our Vis-DSS Framework}

\subsection{Our Engine}
In order to make our toolkit easy to incorporate in applications, we assemble a summarization engine which is used to solve our tasks of image and video summarization, training data subset selection and diversified active learning.

We provide easy to use APIs and function calls in our library, which allow the user to  effortlessly run the various summarization models present in our toolkit.

The combinations of submodular models and algorithms available in our system are given in Table \ref{tab:summEngine}.

\begin{table}
    \centering
    \resizebox{\columnwidth}{!}{\begin{tabular}{| c | c | c |}
    \hline
         \multirow{3}{1.8cm}{\centering Submodular Function Type} &  \multirow{3}{2cm}{\centering Summarization Algorithm} & \multirow{3}{4cm}{\centering Summarization Model} \\
         & & \\
         & & \\
    \hline
    \multirow{13}{1.8cm}{\centering Similarity Based} & \multirow{5}{2cm}{\centering Budgeted Greedy} & Disparity Min\\
    \cline{3-3}
    & & Max Marginal Relevance\\
    \cline{3-3}
    & & Facility Location\\
    \cline{3-3}
    & & Graph Cut\\
    \cline{3-3}
    & & Saturated Coverage\\
    \cline{2-3}
         & \multirow{5}{2cm}{\centering Stream Greedy} & Disparity Min\\
    \cline{3-3}
    & & Max Marginal Relevance\\
    \cline{3-3}
    & & Facility Location\\
    \cline{3-3}
    & & Graph Cut\\
    \cline{3-3}
    & & Saturated Coverage\\
    \cline{2-3}
         & \multirow{3}{2cm}{\centering Coverage Greedy} & Facility Location\\
    \cline{3-3}
    & & Graph Cut\\
    \cline{3-3}
    & & Saturated Coverage\\
    \hline
    \multirow{6}{1.8cm}{\centering Coverage Based} & \multirow{3}{2cm}{\centering Budgeted Greedy} & Set Cover\\
    \cline{3-3}
    & & Probabilistic Set Cover \\
    \cline{3-3}
    & & Feature Based Functions\\
    \cline{2-3}
         & \multirow{3}{2cm}{\centering Coverage Greedy} & Set Cover\\
    \cline{3-3}
    & & Probabilistic Set Cover \\
    \cline{3-3}
    & & Feature Based Functions\\
    \hline
    \end{tabular}}
    \caption{Combination of Submodular Models and Algorithms available in Vis-DSS.}
    \label{tab:summEngine}
\end{table}

\subsection{Image Collection Summarization}
Image Collections, such as albums of vacations photographs, generally contain a large number of repetitive and similar looking images. Humans are able to identify this redundancy intuitively, however it may be a taxing process due to the sheer number of images in a large image collection. To summarize such image collections, we provide the following variants of summarization in our toolkit.

\begin{figure}
\centering
    \includegraphics[width=0.45\textwidth]{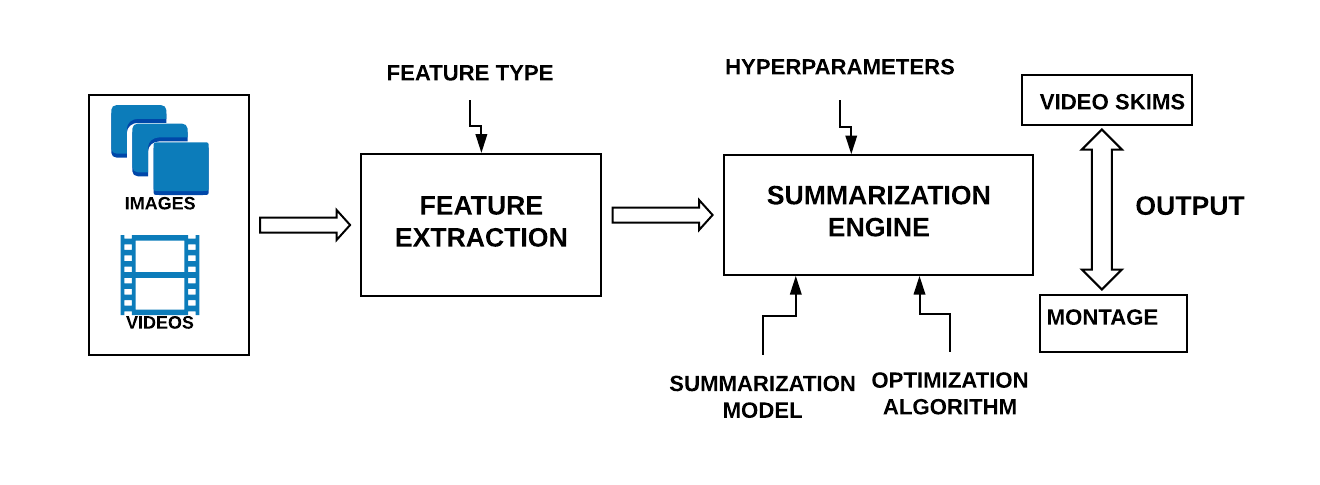}
    \caption{Flow Chart of Steps Involved in Extractive Summarization}
    \label{fig:extractSumm}
\end{figure}

\subsubsection{Extractive Summarization}
Extractive Summarization extracts a set of diverse, yet representative frames from the original image collection. The summary is determined at run-time based on the desired length, or the desired coverage, representation and diversity. The output summary can be in the form of a stitched video of summary images or an image montage.

\paragraph{Feature Extraction}
In order to obtain a diverse subset from the collection, we first need to extract features for every image in the collection. These features would then be used by the submodular summarization models to select the elements to be added to the summary set.

Two types of features can be extracted for an image in our system. They are as follows:

\begin{enumerate}\label{featExtract}
    \item \emph{Simple Feature}: The color histogram is computed for the Hue and Saturation channels of the image. This color histogram is then normalized and it acts as the feature for the image.
    \item \emph{Deep Feature}: We use CNN models to extract features from their fully connected layers. These features are high-dimensional and are used by the summarization model to create the summary set.
\end{enumerate}

While performing Deep Feature Extractive Summarization in our system, we extract the features from the first fully connected layer of the 2012 Imagenet winning model, AlexNet by \citeauthor{AlexNet} \cite{AlexNet}. We obtain a 4096 dimensional feature vector for every image, which is then fed to the summarization model.

\begin{figure}
\centering
    \includegraphics[width=0.45\textwidth]{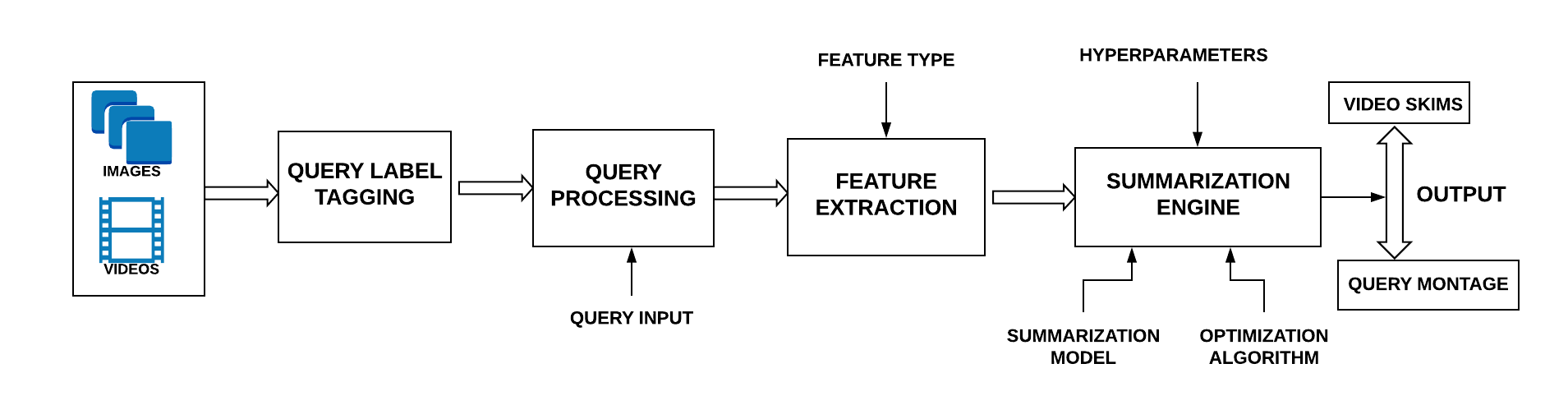}
    \caption{Flow Chart of Steps Involved in Query Summarization}
    \label{fig:querySumm}
\end{figure}

\subsubsection{Query Summarization}
Given an input query, extract a set of diverse and representative frames relevant to that query. Once the relevant frames have been retrieved, the procedure is similar to extractive summarization. The output is a stitched video or image montage of images apropos to the query.

\paragraph{Query Label Tagging}
The first step in Query Summarization involves the assignment of a label to every image in the collection. A user can then input a query based on these labels. In order to tag the images, we use a scene classification CNN model \cite{zhou2014learning} to classify the background scenes in the images and assign the corresponding label to it. 

\paragraph{Query Processing}
Once all the images in the collection are labelled, an input query is accepted from the user which is used to select the appropriate frames from the collection.

\paragraph{Feature Extraction}
As used in extractive summarization, the AlexNet model is used to extract deep features for all the selected relevant frames of the collection.

\subsubsection{Entity Summarization} \label{imageEntitySumm}
An image collection may contain multiple occurrences of numerous distinct entities. These entities can be of different types, such as person, vehicle, animals, etc. Entity Summarization identifies and extracts these entities from the input and summarizes them to return an image montage consisting of diverse entities present in it. The same entity summarization approach is also used to identify and output a representative set of faces present in our input collection.

\paragraph{Detecting and Extracting Entities}
To detect the different entities present in the input video, we use a YOLOv2 \cite{YOLO9000} object detection model trained on the PASCAL VOC \cite{VOC} dataset. The model is capable of detecting objects belonging to various categories, such as person, vehicles, animals and indoor objects. This allows us to identify common objects seen in everyday scenarios and extract them, in order to perform entity summarization on them.

Another variant of Entity Summarization uses a Single Shot Detector \cite{SSD} based ResNet \cite{Resnet} model trained on human faces. The model detects all faces present in the image collection. We then perform summarization on this collection of extracted faces to identify the different people present in the input image collection.

\paragraph{Feature Extraction}
Similar to the other variants of summarization, we continue using CNN models to extract features of our extracted entities. The AlexNet model is used to extract features of the entities detected by the YOLOv2 model, whereas a Deep Metric Learning based ResNet model \cite{dlibFace} is used for obtaining features when the extracted entities are human faces.

\subsection{Video Summarization}
Similar to Image Collection Summarization, our system is also capable of summarizing videos by identifying the diverse, representative or interesting snippets in a video. We provide the same variants of summarization for video summarization, as provided for Image Collection summarization, i.e. extractive summarization and query summarization, along with the addition of entity summarization.

\paragraph{Snippet Types}
To summarize a video, we first create snippets of the video. These snippets form our ground set, which is used by our summarization models to generate the summary.

Two kinds of snippets can be selected in our system They are as follows:
\begin{enumerate}
    \item \emph{Fixed Length Snippets}: Each element of the ground set is of a fixed snippet size, which can be configured by the user. By default, we set the snippet size to two seconds.
    \item \emph{Shot Detected Snippets}: Instead of having snippets of fixed size, we provide the option of having individual shots as snippets. This ensures that the snippets do not start/end abruptly, and the generated summary video looks uniform and smooth.
\end{enumerate}

\subsubsection{Extractive Summarization}
Similar to extractive summarization in Image Collection Summarization, we extract a set of diverse, yet representative snippets from the input video. The summary can be generated in form of image montage or video skim of the original video.

\paragraph{Feature Extraction} \label{vidFeatExtract}
As in the case of Image Collection Summarization, the first step in Extractive Video Summarization is feature extraction. We use the same types of features for video summarization as described in section \ref{featExtract}. However, in the case of video summarization, we extract these features for all frames in a given snippet and create an average feature vector representing that snippet.

\subsubsection{Query Summarization}
In line with query summarization for Image Collection Summarization, we extract a set of diverse and representative snippets relevant to the input query.

\paragraph{Query Label Tagging}
For every snippet in the ground set, we pass all frames through the scene classification CNN model \cite{zhou2014learning} and tag the individual frames in the snippet.

\paragraph{Query Processing}
After tagging all snippets, an input query is accepted from the user. If a snippet contains a frame which satisfies the input query, the snippet is retained in the ground set, else it is removed from the ground set.

\paragraph{Feature Extraction}
To perform summarization on the query relevant snippets, we follow the same methodology as followed in section \ref{vidFeatExtract}. We extract deep features of every frame in a particular snippet and aggregate those features, so as to generate an average feature vector which is representative of that snippet.

\begin{figure}
\centering
    \includegraphics[width=0.45\textwidth]{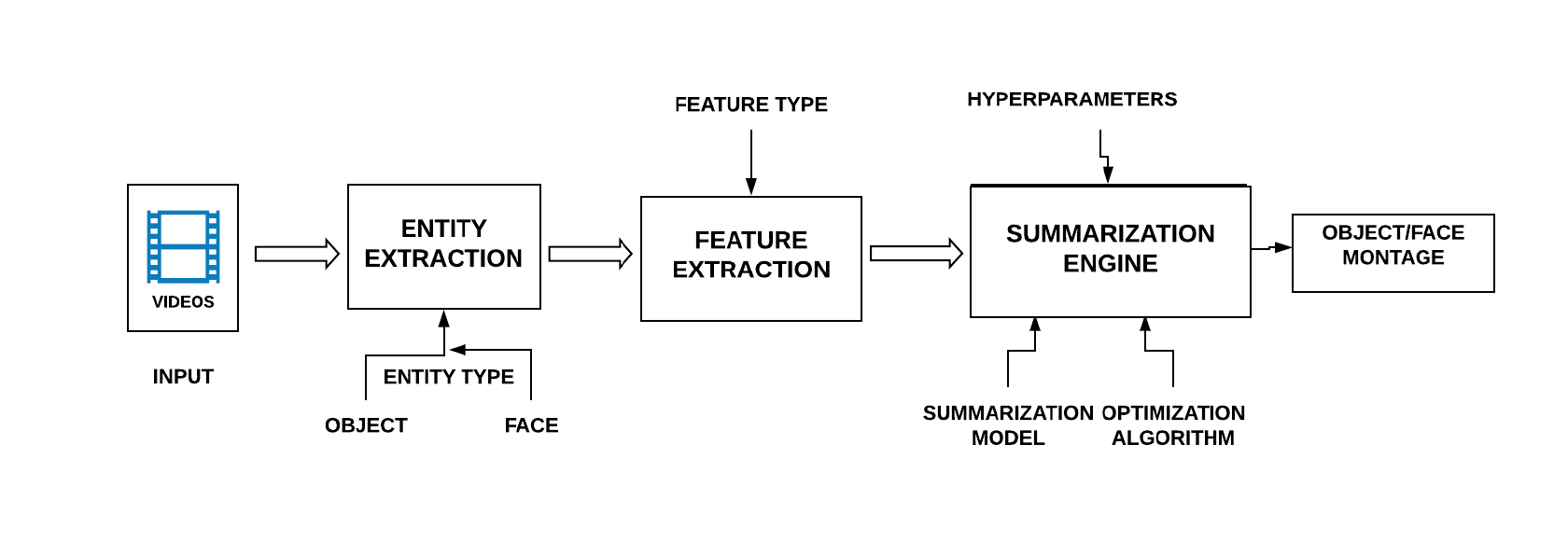}
    \caption{Flow Chart of Steps Involved in Entity Summarization}
    \label{fig:entitySumm}
\end{figure}

\subsubsection{Entity Summarization}
Similar to Image Collection Summarization, videos may also contain multiple occurrences of numerous distinct entities and faces. We adopt the same approach that we've taken in section \ref{imageEntitySumm}.

\paragraph{Detecting and Extracting Entities}
To detect the different entities present in the input video, we use the same YOLOv2 \cite{YOLO9000} object detection model to detect common place objects, and a SSD based ResNet face detector to extract our required entities.

\paragraph{Feature Extraction}
In case of entity summarization, we follow the same feature extraction pipeline used in the Image Collection Summarization case, which has been elaborated upon in section \ref{imageEntitySumm}.

\begin{figure}
 \centering
     \includegraphics[width=\linewidth]{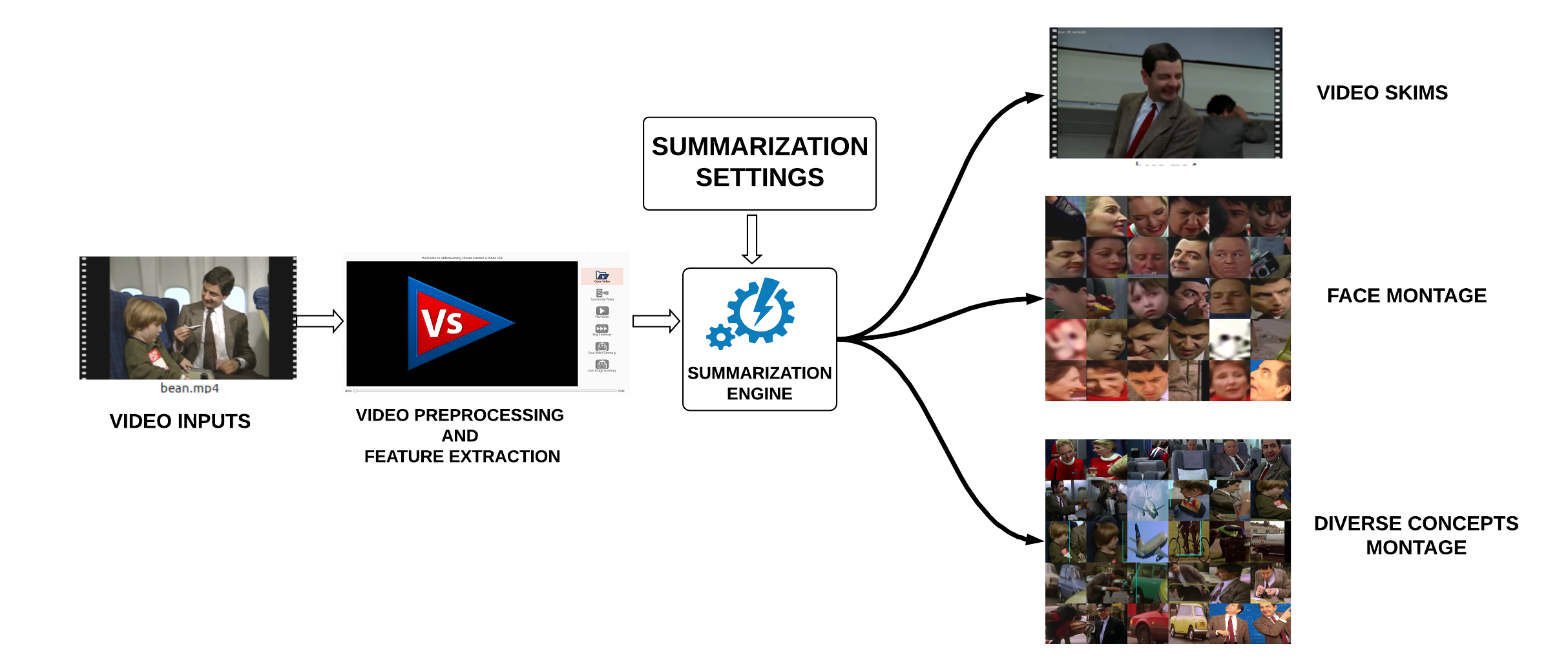}
     \caption{Showing the application workflow through an example video.}
     \label{fig:appflow}
\end{figure}

\begin{figure*}
    \centering
    \includegraphics[width = 0.32\textwidth]{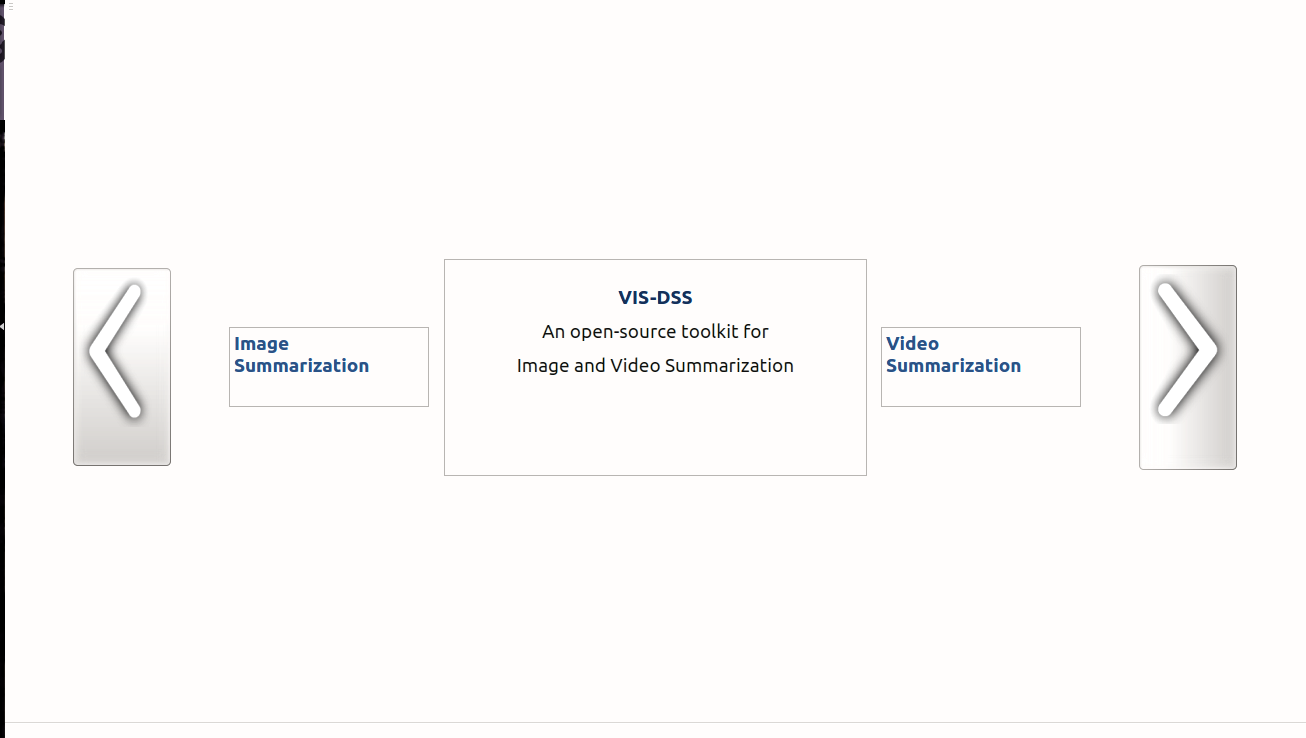}
    ~
    \includegraphics[width = 0.32\textwidth]{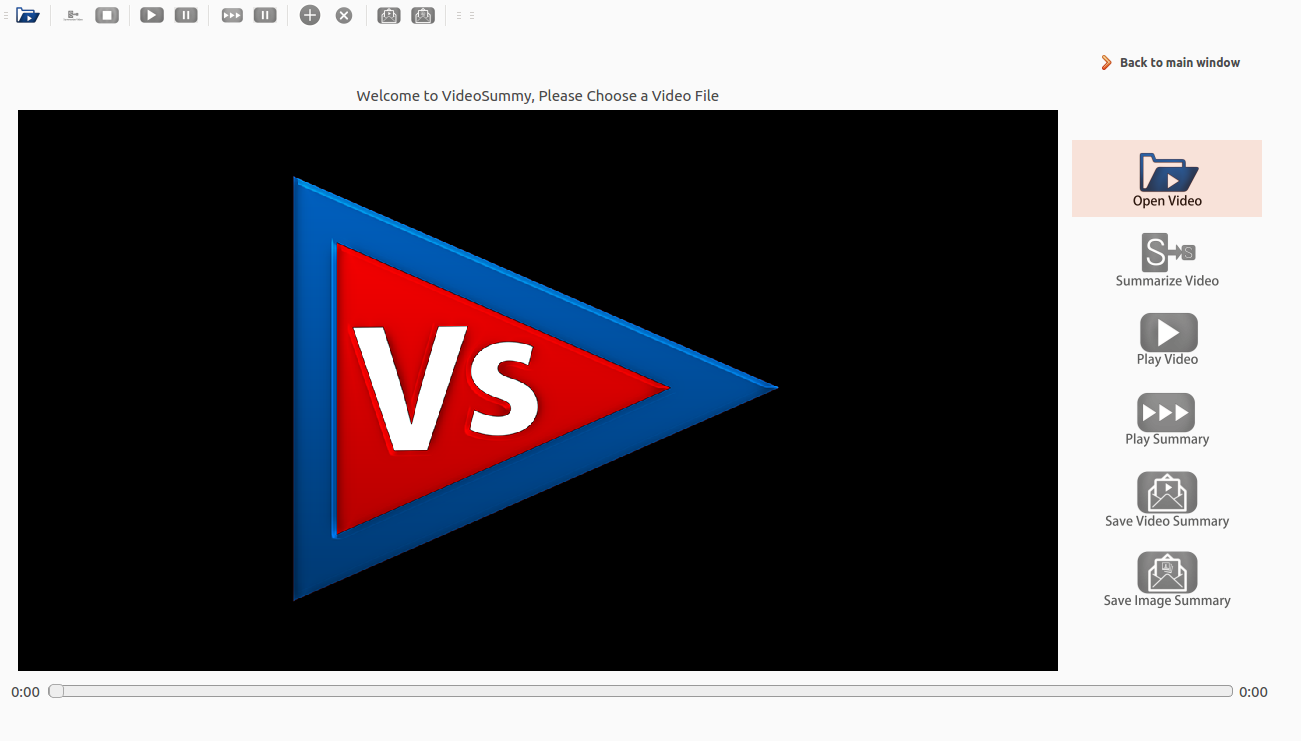}
    ~
    \includegraphics[width = 0.32\textwidth]{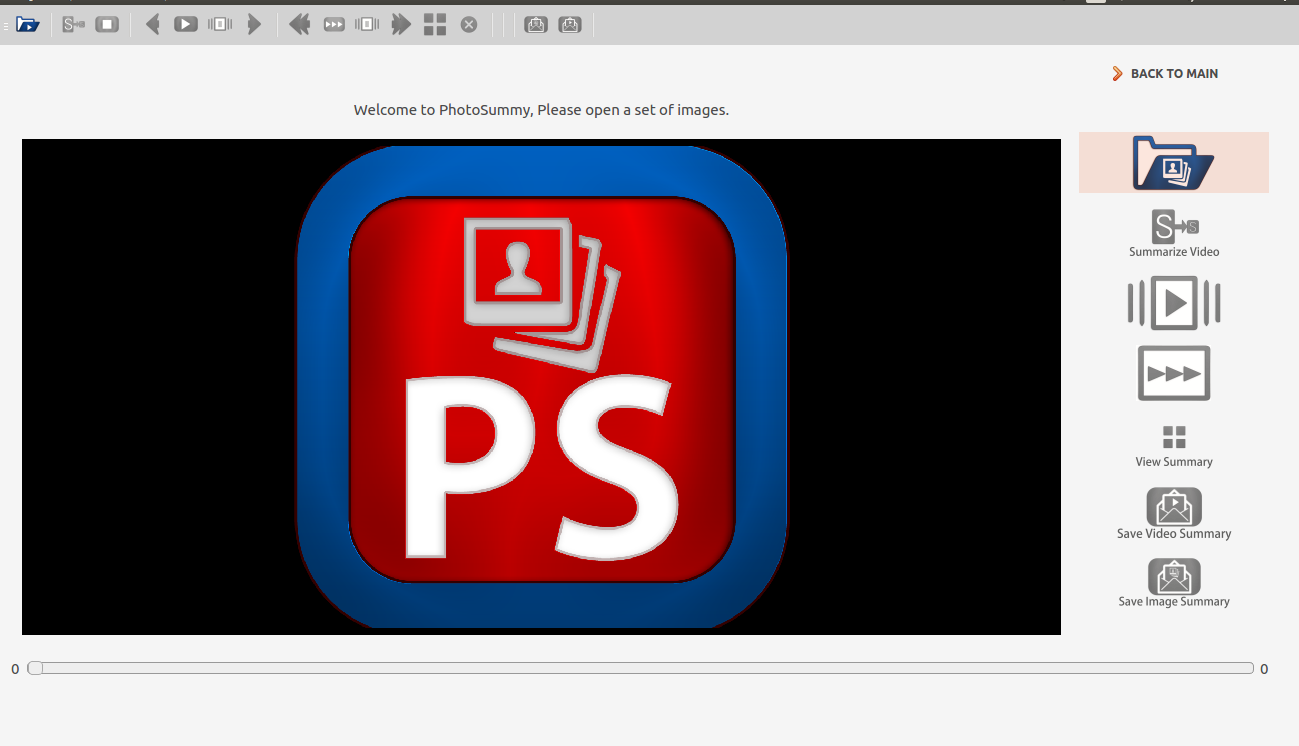}
    ~
    \includegraphics[width = 0.32\textwidth]{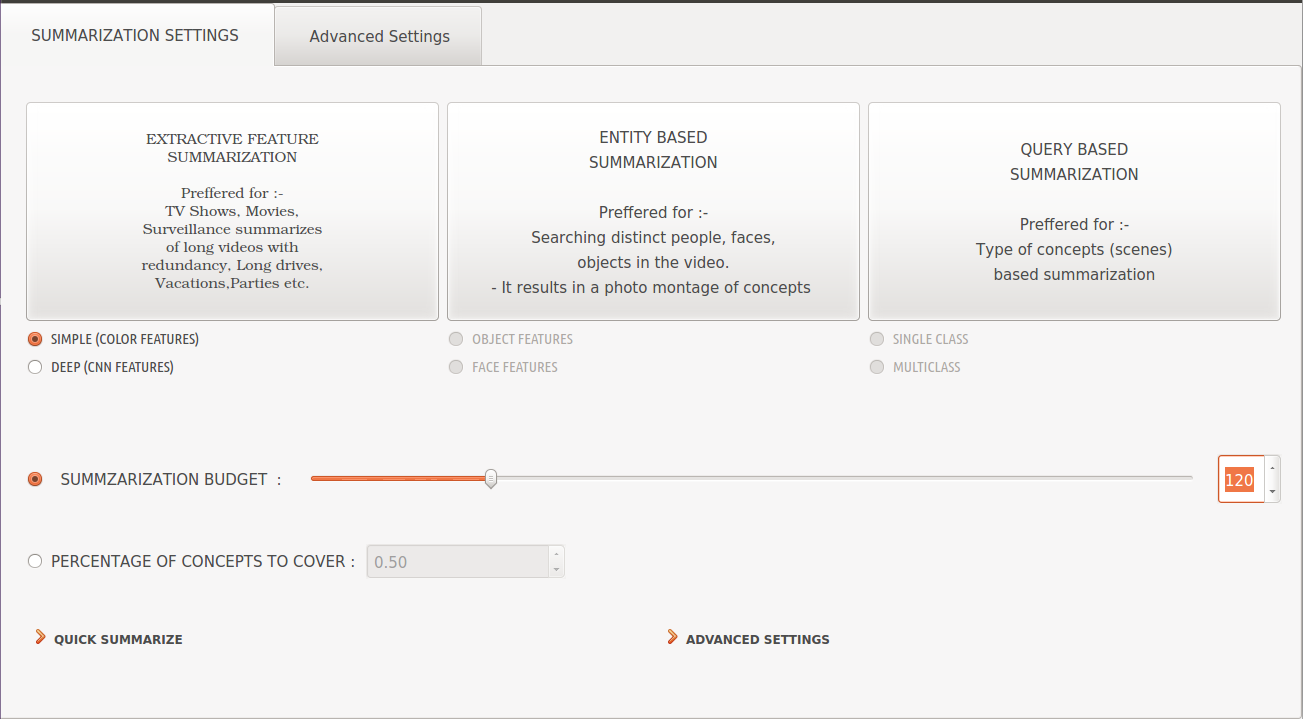}
    ~
    \includegraphics[width = 0.32\textwidth]{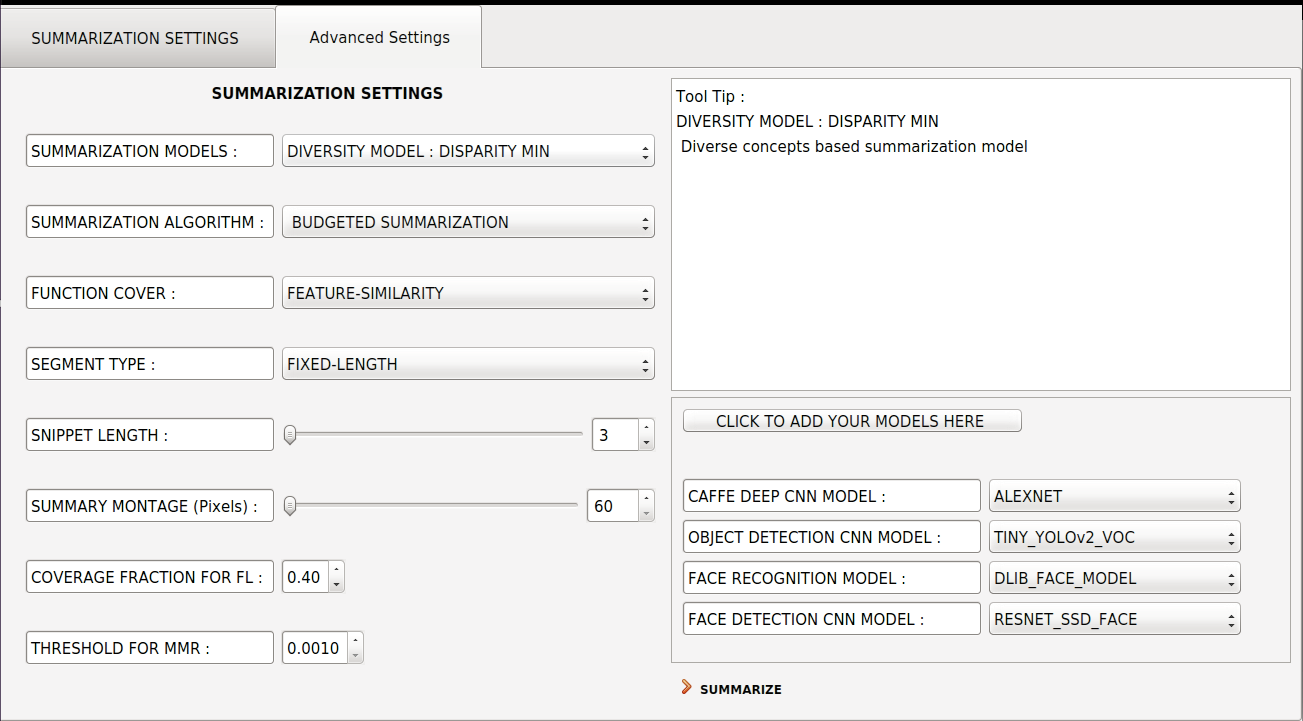}
    \caption{The different pages of the GUI application. First page allows us to choose between Video or Image summarization, the second and third open up the Video and Image Summarization pages (where you can load the video or image collection). Finally, the last two pages show the advanced settings for choosing the different summarization models, features etc.}
    \label{fig:app-screenshots}
\end{figure*}

\begin{figure}
\centering
    \includegraphics[width=0.7\columnwidth]{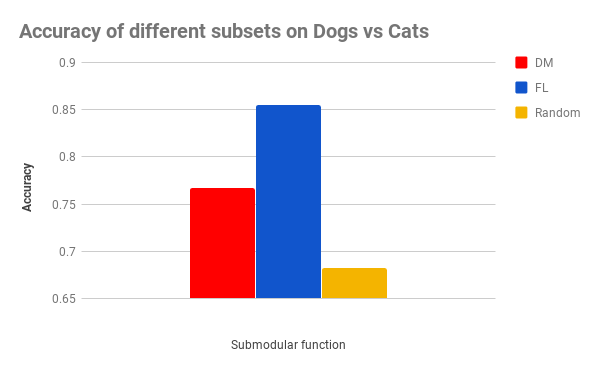}
    ~
    \includegraphics[width=0.7\columnwidth]{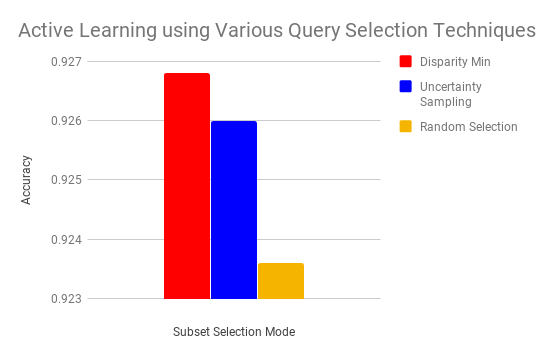}
    ~
    \includegraphics[width=0.7\columnwidth]{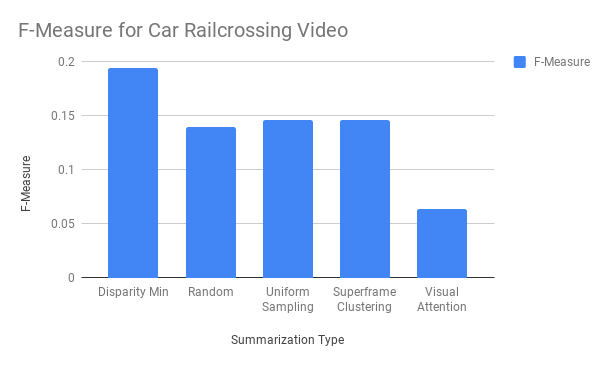}
    \caption{Comparison of submodular summarization for data subset selection, diversified active learning and video summarization}
    \label{fig:dss}
\end{figure}

\begin{figure}
\includegraphics[width=0.5\textwidth, clip=true]{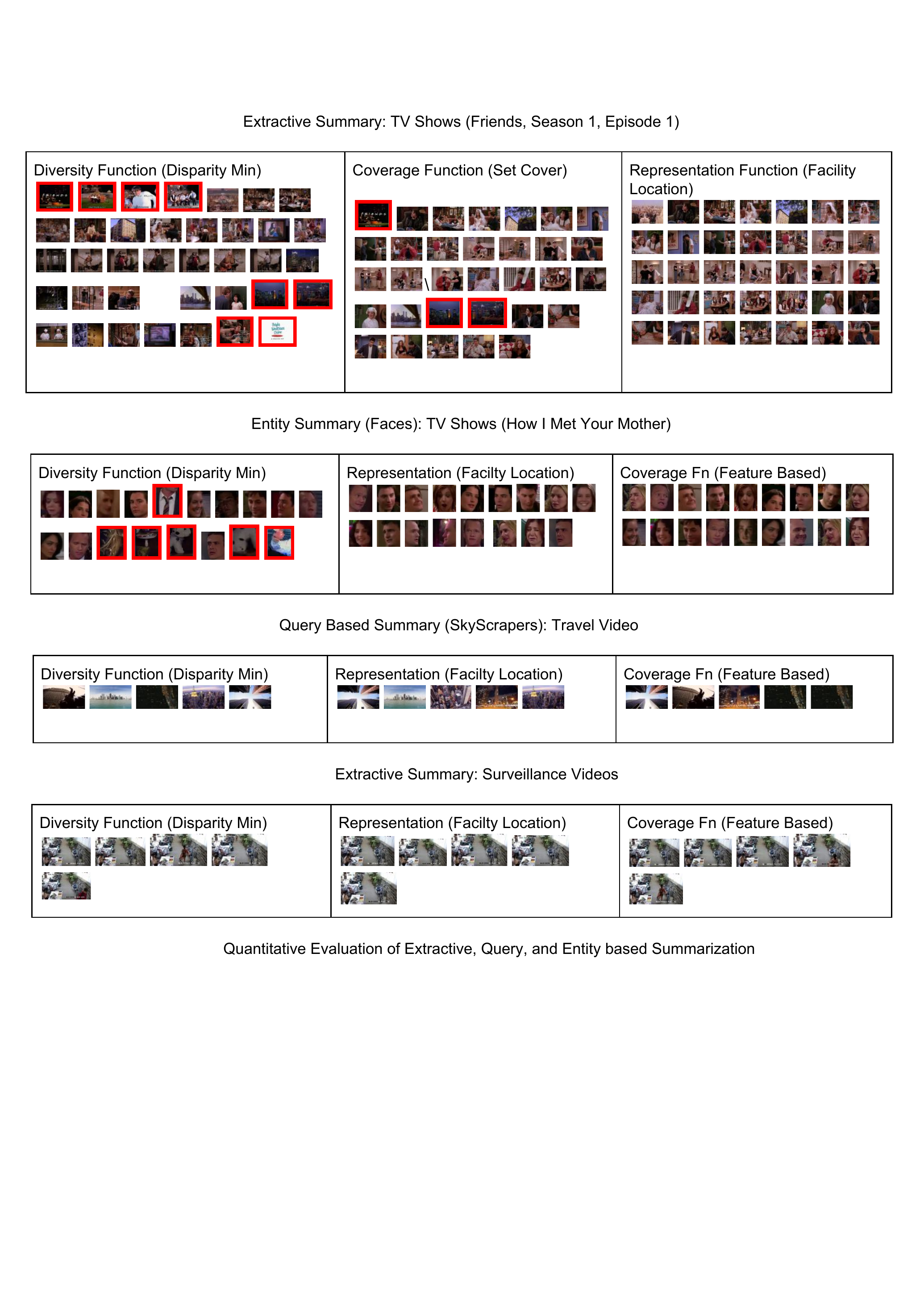}
\vspace{-3.3cm}
\caption{Illustration of the Results. The top figure shows the results from extractive summarization on TV shows, the second demonstrates entity summary on a TV show. The third figure shows the results of query based summarization on a query "SkyScraper" while the fourth one shows the results of extractive summarization on surveillance videos. In each case we compare Representation, Diversity and Coverage models. See the text for more details.}
\label{results}
\end{figure}

\begin{figure}
\centering{
\includegraphics[width=0.21\textwidth]{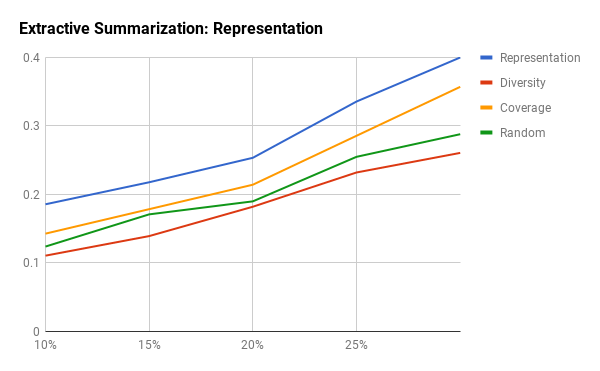}
~
\includegraphics[width=0.21\textwidth]{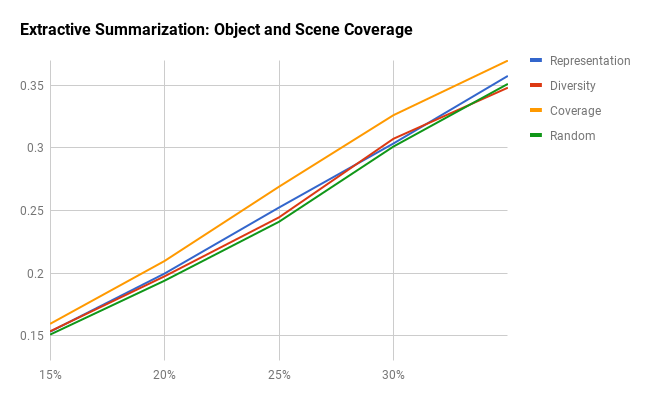}

\includegraphics[width=0.21\textwidth]{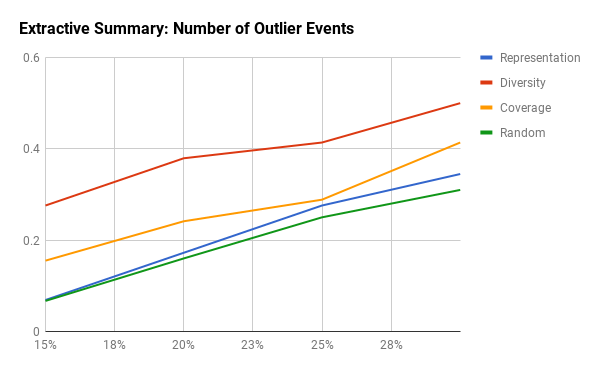}
~
\includegraphics[width=0.21\textwidth]{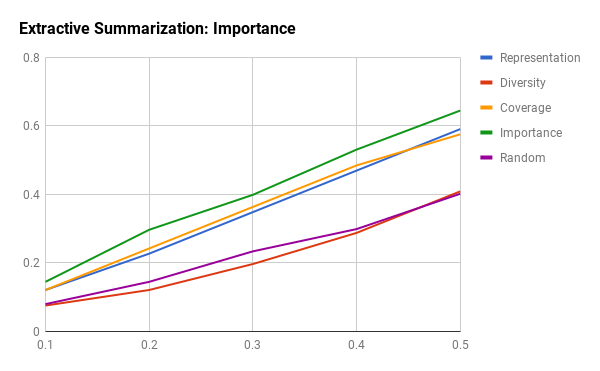} 

\includegraphics[width=0.21\textwidth]{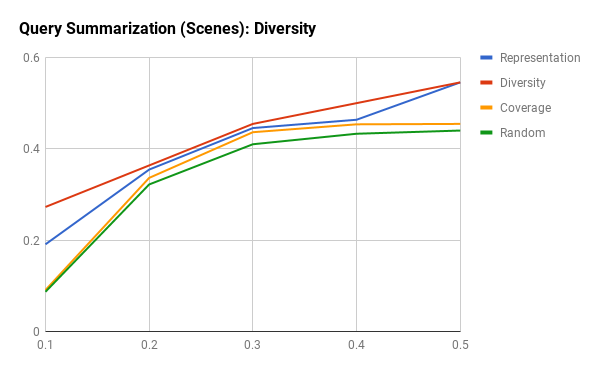}
~
\includegraphics[width=0.21\textwidth]{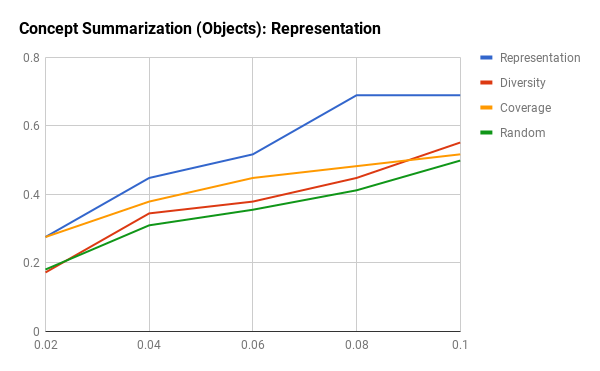}}
\caption{Comparison of Diversity, Coverage and Representation Models for various domains and scenarios. See the text for more details.}
\label{quantres}
\end{figure}

\subsection{Visual Training Data Subset Selection}
The goal behind data subset selection is to select a small representative dataset which can be used for training models efficiently, performing quick hyperparameter tuning or reducing the data labeling cost. A subset selected using diversity models for data summarization helps towards this goal, as opposed to selecting a random subset of the same size. This technique can be used for a wide variety of computer vision tasks which can be solved using deep CNNs such as image classification and object detection. 

Given a dataset for an image classification task, we can use Vis-DSS to generate a subset of that dataset using a given submodular function to train a model without a significant loss in accuracy and show that this subset outperforms a random subset of the same size. The features used to compute the representativeness for subset selection can be AlexNet features trained on ImageNet, GoogLeNet trained on the Places205 dataset \cite{zhou2014learning} or any such pre-trained model. 

We demonstrate using such a subset using two submodular functions, Disparity Min and Facility Location to train an AlexNet model on the Dogs vs Cats dataset \cite{CatsvsDogs} and compare the accuracy with a random subset. 

\subsection{Diversified Active Learning}
Since submodular functions naturally model diversity, they are a perfect fit for selecting samples to be labeled in order to make a learning algorithm better. A blend of subset selection algorithms with active learning can speed up the learning process and maximize gradient flow in the very initial training epochs. We propose diversified active learning that uses various submodular functions that act as query selection techniques for datasets of all types. Deep learning being a highly iterative process, it is imperative to get insights regarding the learning of a model during its early training epochs.

Vis-DSS is the solution to all the query selection problems for active learning. Depending on requirements based on representation, diversity or coverage needed from a particular domain, several sudmodular functions are provided in Vis-DSS to aid the selection of the best query from a dataset. This is an iterative process and required features that would be diverse in nature. For instance, the features could be VGG or AlexNet features trained on ImageNet, or YOLO features trained on COCO.

We illustrate the performance of different summarization models for data subset selection and active learning. We used the Dogs vs Cats dataset \cite{CatsvsDogs}, as an example to illustrate this. The features are extracted using AlexNet which is pre-trained on ImageNet.

\section{Applications built on Vis-DSS}
\subsection{The Vis-DSS Summarization Application}
 Vis-DSS is an open-source standalone application which provides implementation of submodular algorithms for visual data subset selection and summarization.It includes a low level optimization engine written in ANSI C++, which is invoked by the graphical user interface (GUI) provided by a multi-threaded application built over QT Application development framework. The Application provides a clean and simple graphical interface through which a user can select an algorithm among different permutations of summarization techniques, which suits requirements the best. As the source has been released under open-source paradigm, it is possible to integrate it to other existing C++ projects. To make that system is easy to understand and user friendly we have have a tool tip inside the application which helps the users to understand different algorithms while selecting among them.
In terms of system dependencies and deployment, Vis-DSS requires OpenCV \cite{opencv_library} for all the image/video manipulations and color feature extractions. Apart from this, the other dependencies are Caffe \cite{jia2014caffe}, for deep feature extractions and query labelling, and Dlib \cite{dlibFace}, for face features. The GUI application has been designed in the Open-Qt Framework, which enables it to be distributed easily as a binary. Figure~\ref{fig:appflow} shows the flow of the application, while Figure~\ref{fig:app-screenshots} shows the screen shots of our application and the different pages.

\section{Experimental Evaluation}

Figure~\ref{fig:dss} (top) demonstrates the results of Data subset selection with submodular functions. We use two specific summarization models, Facility Location and Disparity Min (one is a representation function and another is diversity). We compare a random baseline. We see that the summarization models consistently outperform the random baseline. Next, we study the performance of the summarization models in the context diversified active learning (Figure~\ref{fig:dss} middle). We use uncertainty sampling and random selection as baselines. We then consider Disparity Min as a diversity model within the context of diversified active learning. Again, we see that disparity min significantly out performs these baselines. 
We also compare our video summarization models on Summe dataset~\cite{GygliECCV14}. Again, we choose Disparity Min as an example summarization model and compare to random, uniform, superframe culstering and visual attention on the Car Railcrossing video. We see that Disparity Min outperforms all baselines in terms of F1 measure.

We next qualitatively and quantitatively study the role of different submodular functions for video/image summarization. Figure~\ref{results} shows the results for extractive summarization, entity summarization and query based summarization on frames from different kinds of videos. We compare the different summarization models under various scenarios and evaluation measures. Instead of comparing all the submodular functions described above, we consider representatives from each class of functions. We use Facility Location as a representative function, Disparity Min for Diversity and Set Cover as a choice for Coverage functions. Towards this, we create a dataset of videos from different categories. We select 10 videos from the following categories: Movies/TV shows, Surveillance camera footage, Travel videos and sports videos like Soccer. In the following sections, we annotate various events of interest (ground-truth) from these videos to define various evaluation criteria. The goal of this is to demonstrate the role of various summarization models. 

\paragraph{Extractive Summarization: Representation} The top Figure in Fig.~\ref{results} demonstrates the results of extractive summarization on Movies and TV shows. Diversity Models tend to pick up outlier events, which in this case, include transition scenes and other outliers. In contrast, the Representation function (Facility Location) tends to pick the representative scenes. The coverage function does something in between. In the case of a TV show, representative shots are probably more important compared to the transition scenes. To quantify this, define an evaluation measure as follows. We divide a movie (TV Show) into a set of scenes $S_1, \cdots, S_k$ where each scene $S_i$ is a continuous shot of events. We do not include the outliers (we define outliers as shots less than a certain length -- for example transition scenes). Given a summary $X$, define $R(X) = \sum_{i = 1}^k \min(|X \cap S_i|, 1)/k$. A summary with a large value of $R(X)$ will not include the outliers and will pick only single representatives from each scene. We evaluate this on $10$ different TV show and movie videos. Figure~\ref{quantres} (top left) compares the representative, diversity, and coverage models and a random summary baseline. We see the representative model (Facility Location) tends to perform the best as expected, followed by the coverage model. The diversity model does poorly since it picks a lot of outliers. 

\paragraph{Extractive Summarization: Coverage} Next, we define an evaluation criteria capturing coverage. For each frame in the video (sampled at 1FPS), define a set of concepts covered $\mathcal U$. Denote $\mathcal U(X)$ as the set of concepts covered by a set $X$. For each frame of the video, we hand pick a set of concepts (scenes and objects contained in the video). Define the coverage objective as $C(X) = \mathcal U(X)/\mathcal U(V)$. Figure~\ref{quantres} demonstrates the coverage objective for the different models. We obtain this by creating a set of 10 labeled videos of different categories (surveillance, TV shows/movies, and travel videos). As expected, the coverage function (set cover) achieves superior results compared to the other models. 

\paragraph{Extractive Summarization: Outliers and Diversity} In the above paragraphs, we define two complementary evaluation criteria, one which captures representation and another which measures coverage. We argue how, for example, representation is important in Movies and TV shows. We now demonstrate how the diversity models tend to select outliers and anomalies. To demonstrate this, we select a set of surveillance videos. Most of our videos have repetitive events like no activity or people sitting/working. Given this, we mark all the different events (what we call outliers), including for example, people walking in the range of the camera or any other different activity. We create a dataset of 10 surveillance videos with different scenarios. Most of these videos have less activity. Given a set $S_1, S_2, \cdots S_k$ of these events marked in the video, define $D(X) = \sum_{i = 1}^k \min(|X \cap S_i|, 1)$. Note this measure is similar to the representative evaluation criteria ($R(X))$ except that it is defined w.r.t the outlier events. Figure~\ref{quantres} (middle left) shows the comparison of the performance of different models on this dataset. As expected, the Diversity measures outperforms the other models consistently.

\paragraph{Extractive Summarization: Importance} To demonstrate the benefit of having the right \emph{importance} or \emph{relevance} terms, we take a set of videos where intuitively the relevance term should matter a lot. Examples include sports videos like Soccer. To demonstrate this, we train a model to predict important events of the video (e.g. the goals, red card). We then define a simple Modular function where the score is the output of the classifier. We then test this out and compare the importance model to other summarization models. The results are shown in Figure~\ref{quantres} (middle right). As we expect, the model with the importance gets the highest scores.

\paragraph{Query Summarization: Diversity} We next look at query based summarization. The goal of query based summarization is to obtain a summary set of frames which satisfy a given query criteria. Figure~\ref{results} (third row) qualitatively shows the results for the query "Sky Scrapers". The Diversity measure is able to obtain a diversity of the different scenes. Even if there is an over-representation of a certain scene in the set of images satisfying the query, the diversity measure tends to pick a diverse set of frames. The representation measure however, tends to focus on the representative frames and can pick more than one image in the summary from scenes which have an over-representation in the query set. We see this Figure~\ref{results}. To quantify this, we define a measure $M(X)$ by dividing the video into a set of clusters of frames $S_1, \cdots, S_k$ where each cluster contains similar frames. These are often a set of continuous frames in the video. We evaluate this on a set of 10 travel videos, and compare the different models. We see that the diversity and representation models tend to perform the best (Figure~\ref{quantres}, bottom left), with the diversity model slightly outperforming the representative models. We also observe that there are generally very few outliers in the case of query based summarization, which is another reason why the diversity model tends to perform well.

\paragraph{Entity Summarization: } Lastly we look at Entity summarization. The goal here is to obtain a summary of the entities (faces, objects, humans) in the video. Figure~\ref{results} (second row) demonstrates the results for Entity summarization of Faces. We see the results for Diversity, Coverage and Representation Models. The diversity model tends to pick up outliers, many of which are false positives (i.e. not faces). The representation model skips all outliers and tends to pick representative faces. To quantitavely evaluate this, we define a representation measure as follows. We remove all the outliers, and cluster the set of entities (objects, faces) into a set of clusters $E_1, \cdots, E_k$ where $E_i$ is a cluster of similar entities. We evaluate this again on a set of 10 videos. Figure~\ref{quantres} (bottom right) shows the results for objects. The results for Faces is similar and in the interest of space, we do not include these. We see that the representation model tends to outperform the other models and skips all the outliers. The diversity model focuses on outliers and hence does not perform as well.

Finally, we demonstrate the computational scalability of our framework. Table 3 shows the results of the time taken for Summarization for a two hour video (in seconds) with and without memoization. The groundset size is $|V| = 7200$. We see huge gains from using memoization compared to just computing the gains using the Oracle models of the functions. All our experiments were performed on Intel(R) Xeon(R) CPU E5-2603 v3 @1.6 GHz (Dual CPU) with 32 GB RAM. Table 4 compares the implementations of Gygli et al~\cite{gyglim} and SFO~\cite{krause2010sfo}. We see substantial gains with our implementations thanks to memoization.
\begin{table}
    \centering
    \begin{tabular}{ |c|c|c|c|c|c|c|}
        \hline
         & \multicolumn{3}{c|}{\textbf{Memoization}} & \multicolumn{3}{c|}{\textbf{No Memoization}} \\
        \hline
        \textbf{Function} & \textbf{5\%} & \textbf{15\%} & \textbf{30\%} & \textbf{5\%} & \textbf{15\%} & \textbf{30\%}  \\
         \hline
        Fac Loc & 0.34 & 0.4 & 0.71 & 48 & 168 & 270  \\
         \hline
        Sat Cov & 0.36 & 0.64 & 0.92 & 55 & 177 & 301  \\
        \hline
         Gr Cut & 0.39 & 0.52 & 0.82 & 41 & 161 & 355  \\
        \hline
        Feat B & 0.16 & 0.21 & 0.32 & 9 & 16 & 21  \\
        \hline
        Set Cov & 0.21 & 0.31 & 0.41 & 5 & 16 & 31  \\
        \hline
        PSC & 0.11 & 0.37 & 0.42 & 7 & 19 & 35  \\
        \hline
        DM & 0.11 & 0.61 & 0.82 & 21 & 125 & 221  \\
                \hline
        DS & 0.21 & 0.63 & 0.89 & 41 & 134 & 246  \\
        \hline
    \end{tabular}
    \caption{Timing results in seconds for summarizing a two hour video for various submodular functions}
    \label{tab:my_label}
\end{table}

\begin{table}
    \centering
    \begin{tabular}{|c|c|c|c|}
       \hline
        Function & Ours & Gygli et al~\cite{gyglim} & SFO~\cite{krause2010sfo}  \\ \hline
        Fac Loc & 0.34 & 26.8 & 52 \\ \hline
        Gr Cut & 0.39 & 35.7 & 43.2 \\ \hline
    \end{tabular}
    \caption{Comparing the running time (in seconds) of our implementation to the code of Gygli et al~\cite{gyglim} and SFO~\cite{krause2010sfo} setting a budget of 5\%.}
\end{table}
\section{Conclusions}
This paper presents Vis-DSS, a unified software framework for Data summarization. We describe our engine based on submodular optimization and provide several applications of it, including Image summarization, video summarization, data subset selection for training and diversified active learning. We also show how we can extend this by building a GUI application for video and image summarization. We hope that this toolkit can be used for applications around visual summarization.

\printbibliography
\end{document}